\def\year{2019}\relax
\documentclass[letterpaper]{article} 
\usepackage{aaai19}  
\usepackage{times}  
\usepackage{helvet}  
\usepackage{courier}  
\usepackage{url}  
\usepackage{graphicx}  
\usepackage{subfigure}
\usepackage{amsfonts,amssymb}
\usepackage{mathrsfs}
\usepackage{amsmath}
\usepackage{booktabs}
\usepackage{multirow}
\usepackage{algorithm}
\usepackage{algpseudocode}

\title{Learning Incremental Triplet Margin for Person Re-identification}
\author{Yingying Zhang, Qiaoyong Zhong, Liang Ma, Di Xie, Shiliang Pu\\
Hikvision Research Institute\\
\{zhangyingying7,zhongqiaoyong,maliang6,xiedi,pushiliang\}@hikvision.com
}

\begin{document}

\maketitle

\begin{abstract}
Person re-identification (ReID) aims to match people across multiple non-overlapping video cameras deployed at different locations. To address this challenging problem, many metric learning approaches have been proposed, among which triplet loss is one of the state-of-the-arts. In this work, we explore the margin between positive and negative pairs of triplets and prove that large margin is beneficial. In particular, we propose a novel multi-stage training strategy which learns incremental triplet margin and improves triplet loss effectively. Multiple levels of feature maps are exploited to make the learned features more discriminative. Besides, we introduce global hard identity searching method to sample hard identities when generating a training batch. Extensive experiments on Market-1501, CUHK03, and DukeMTMC-reID show that our approach yields a performance boost and outperforms most existing state-of-the-art methods.
\end{abstract}

\section{Introduction}\label{introduction}

In recent years, person re-identification (ReID) has aroused concerns of more and more researchers due to its wide range of applications in security and video surveillance. It is a challenging task because of varying illumination conditions, human occlusion, background clutter and different camera views. Most existing methods use a feature vector to represent each person image and then match them with a specific metric. With the emergence of deep learning, feature representations learned with convolutional neural networks (CNN) \cite{krizhevsky2012imagenet,lecun1989backpropagation} immensely outperform hand-crafted features.

\begin{figure}[t!]
  \subfigure[LITM training procedure.]{
    \begin{minipage}[t]{\linewidth}
      \centering
      \includegraphics[width=1\textwidth]{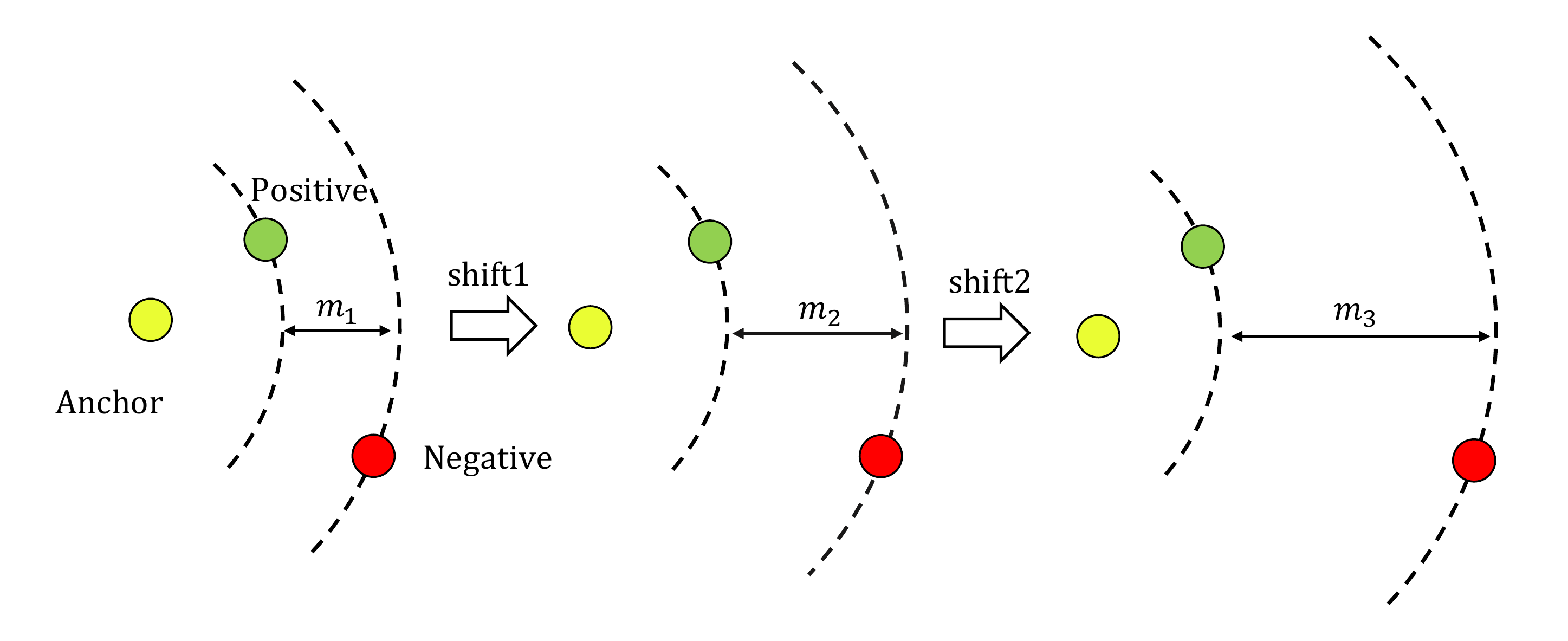}
    \end{minipage}
    \label{e2h-illustration}
  }

  \subfigure[Hard identity examples (yellow: anchor, red: negative).]{
    \begin{minipage}[t]{\linewidth}
      \centering
      \includegraphics[width=1\textwidth]{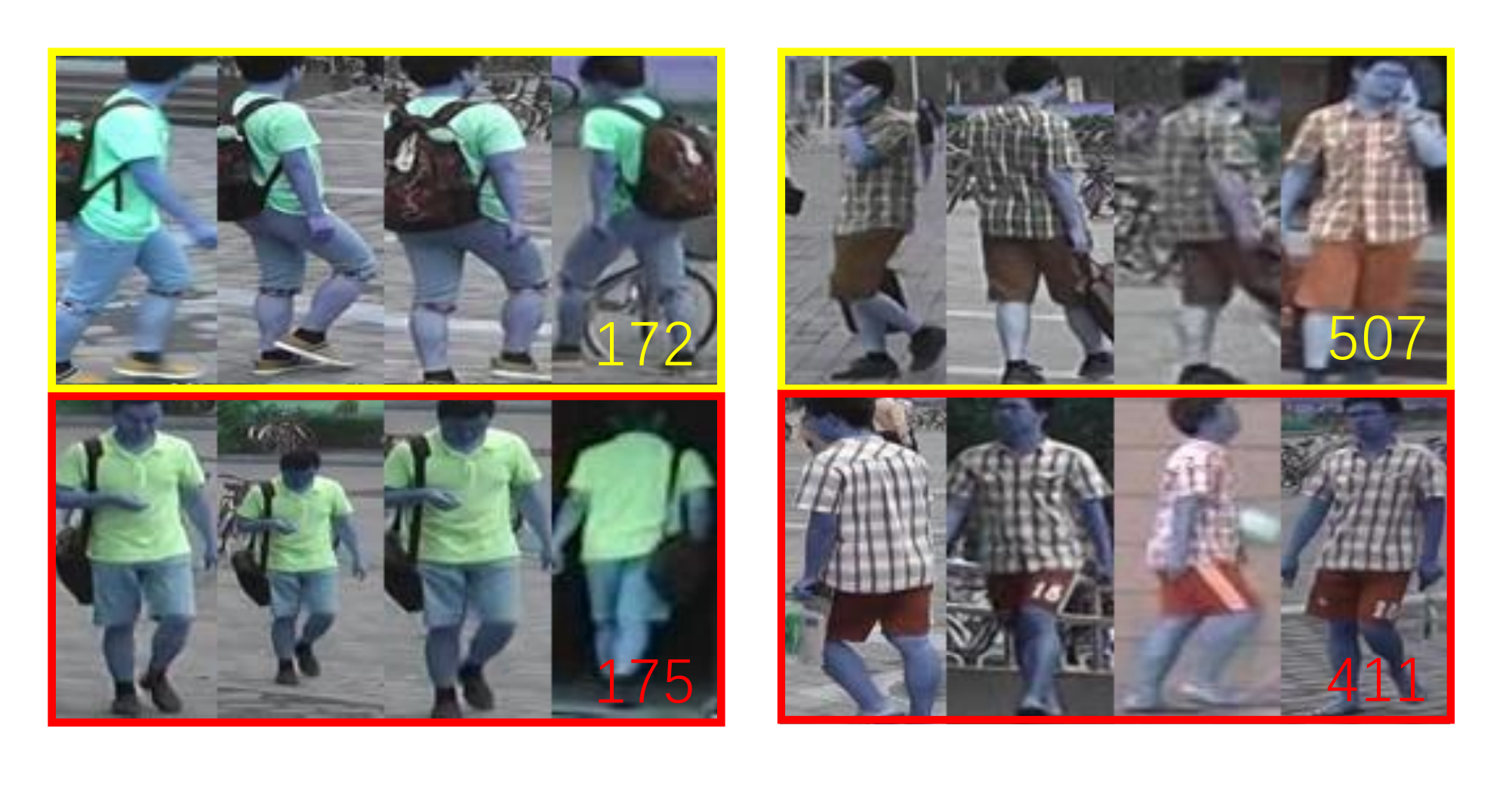}
    \end{minipage}
    \label{hard-identities}
  }
  \caption{(a) illustrates the proposed LITM training procedure. By repeatedly applying a shift on the data points in the embedding space, the margin between the positive and negative pairs in the triplet is progressively increased. (b) shows two examples of hard identity pair on the Market-1501 dataset. The number on the bottom right is the identity label of each person.}
\end{figure}

Currently, the most commonly used loss functions are triplet loss, classification loss and verification loss. Triplet loss was first introduced by \cite{weinberger2009distance}. It directly optimizes a deep CNN which produces embeddings such that positive examples are closer to an anchor example than negative examples. For classification loss, each identity of person in the training set is considered as a class, and the network is trained to classify them correctly. Subsequently, the trained network is used as a feature extractor and a specific metric is chosen to rank the extracted features. In general, the performance of classification loss is superior over triplet loss, since it enforces global inter-class separability in the embedding space. However, as the number of identities increases, the number of learnable parameters grows. For scenarios with very large quantity of identities, it would be non-trivial to train a classification loss. Lastly, verification loss is used to learn a cross-image representation. The network predicts the similarity between two input images directly. During inference, all query-gallery image pairs need to go through the whole network, which is very expensive.

Triplet loss attempts to enforce a margin between the positive and negative pairs of each triplet. Surprisingly, the impact of the margin on ReID performance has not been explored yet in the literature. Intuitively, larger margin leads to better performance. However, as shown in our experiment (Table~\ref{cmp-diff-delta}), simply increasing the margin value of triplet loss does not work well. Instead, we propose a novel training strategy named Learning Incremental Triplet Margin (LITM). As shown in Figure~\ref{e2h-illustration}, we learn a large margin in an multi-stage manner. Firstly, decent positions of the triplet examples in the embedding space are learned using triplet loss with a small base margin. Next, a shift of each data point in the embedding space is learned, which enlarges the gap between the positive and negative pairs. This step is repeatedly applied so that the margin gets increased in an incremental manner. Since mid-level features of the network contain more detailed information and are thus helpful to differentiate identities with similar appearance, we learn the feature shifts using multiple mid-level features. It is worth noting that all the components are implemented in the same network and optimized end-to-end. With LITM, the performance of triplet loss gets significantly improved.

Training of triplet loss requires sampling of triplets from all training images. The number of possible triplets grows cubically with the number of images in the training set. To train triplet loss efficiently, \cite{hermans2017defense} proposed batch hard triplet mining. Firstly, a batch of images is generated by randomly sampling $P$ identities and $K$ images per-identity. Then, for each sample in the batch, the hardest positive and negative samples within the batch are selected to form the triplets. It solves the impractical long training problem partially. However, sampling identities randomly may not ensure that negative pairs are hard enough. For instance, two persons with similar appearance may be dispersed to different batches so that there is no hard negative pair in one batch. To address this issue, we introduce a new identity sampling method called Global Hard Identity Searching (GHIS). We compute the pairwise mean embedding distances of all identities, which measure their dissimilarities. Then identities with small distance (similar appearance) are put together to form a batch. Figure~\ref{hard-identities} shows two examples of searched hard identities from the Market-1501 dataset. We can see that different persons may wear clothes with similar color or texture, which makes them difficult to distinguish even for human beings.

When designing the network architecture for person ReID, it is currently a best practice to adapt from a pre-trained network, e.g. ImageNet pre-trained ResNet-50 \cite{he2016deep}. However, vanilla ResNets were designed for the task of coarse-grained image classification. While person ReID requires a fine-grained recognition within the person category. To narrow the gap, we revisit the ReID problem carefully and propose some guidelines on its network design. Following the guidelines, we make some tweaks to the feature extractor network, which yields a strong baseline implementation of triplet loss.
 
In summary, the main contributions of this paper are as follows:
\begin{itemize}
  \item We propose a novel training strategy which learns incremental triplet margin and leads to significant performance improvement. 
  \item We introduce global hard identity searching method which samples hard identities and makes the training more efficient and effective.
  \item We propose some guidelines on network design for the task of person ReID and yield a strong triplet loss baseline.
  \item Combining all the improvements, we achieve state-of-the-art performances on common person ReID benchmarks.
\end{itemize}

\section{Related Work}\label{related-work} 

\textbf{Person ReID}~~~For the task of person ReID, most existing approaches attempt to learn identity-discriminative representation of person images with supervised learning. With the recent advancements of deep learning, this field has been dominated by deep neural networks. \cite{Xiao_2016_CVPR} used a classification loss to learn deep feature representations from multiple domains. \cite{Qian_2017_ICCV} proposed a novel multi-scale deep learning model that is able to learn deep discriminative feature representations at different scales. Their method can automatically determine the most suitable scales for matching. \cite{Shen_2018_CVPR} proposed a Kronecker Product Matching module to generate matching confidence maps between two pedestrian images. \cite{Guo_2018_CVPR} proposed a fully convolutional Siamese network to improve the measurement of similarity between two input images. Rather than feature learning from the whole person image, some other works exploit part-based features. \cite{Yao2017Deep} clustered the coordinates of maximal activations on feature maps to locate several regions of interest. \cite{Zhao_2017_ICCV} embedded the attention mechanism in the network, allowing the model to decide where to focus by itself. In addition, some works attempt to incorporate extra information like human pose and appearance mask to facilitate person ReID. \cite{zheng2015scalable} proposed to extract separate features of different body regions and merge them using a tree-structured fusion network. \cite{Su_2017_ICCV} proposed a pose-driven deep CNN model which explicitly leverages the human part cues to learn effective feature representations.

\noindent\textbf{Triplet Loss}~~~Strictly speaking, triplet loss was first introduced by \cite{weinberger2009distance}. They trained the metric with the goal that the k-nearest neighbors belong to the same class and examples of different classes can be dissociated by a large margin. Based on this work, \cite{schroff2015facenet:} improved the loss to learn a unified embedding for face recognition. They pushed forward the concept of triplet and minimized the distance between an anchor and a positive while maximized the distance between the anchor and a negative. \cite{Cheng_2016_CVPR} improved the triplet loss function by restricting positive pairs within a small distance. And this improved loss was used to train a multi-channel parts-based convolutional neural network model. Recently, \cite{hermans2017defense} summarized the works of ReID using triplet loss, and proposed some training strategies to improve the performance of triplet loss. While our work is also based on triplet loss, we investigate the influence of the margin, which has received little attention so far.

\noindent\textbf{Hard Example Mining}~~~Hard example mining has been widely exploited to assist training of deep neural networks. \cite{Shrivastava_2016_CVPR} proposed online hard example mining to improve the performance of object detection. \cite{hermans2017defense} extended this idea and selected the hardest positive and negative samples within a batch when generating triplets. These methods can be categorized as \emph{local} hard example mining considering that hard examples are mined from a training batch instead of the whole training set. While the proposed GHIS searches hard identities \emph{globally} from all identities in the training set.

\section{Method}\label{method}

Since our approach is based on triplet loss, let us first briefly recap its formulation. Training of triplet loss requires carefully designed sampling of triplets. A triplet consists of an anchor image $x_i^a$, a positive image $x_i^p$ of the same person as the anchor and a negative image $x_i^n$ of a different person. Triplet loss aims to learn a feature embedding so that $x_i^a$ is closer to $x_i^p$ than it is to $x_i^n$ in the embedding space. It can be formulated as follows:
\begin{equation}
  \begin{aligned}
    \mathcal{L}_0 &= \sum_i^N\Big[d^{ap}_0 - d^{an}_0 + m_0\Big]_+, \\
        d^{ap}_0 &= ||f_0(x_i^a) - f_0(x_i^p)||_2^2 \\
        d^{an}_0 &= ||f_0(x_i^a) - f_0(x_i^n)||_2^2 
  \end{aligned}
  \label{l-triplet}
\end{equation}
where $[\cdot]_+$ is the hinge function. $d^{ap}_0$ and $d^{an}_0$ are the squared Euclidean distance between the anchor-positive and anchor-negative pairs respectively. $m_0$ is the margin enforced between $d^{an}_0$ and $d^{ap}_0$. $N$ is the number of triplets in a training batch. $f_0(x_i)\in\mathbb{R}^{d}$ denotes the $d$-dimensional feature embedding of $x_i$. Here we apply a subscript $0$ on $f$, $d$ and $m$ to indicate the base feature vector, distance and margin respectively. They will be updated later.

\subsection{Learning Incremental Triplet Margin}
To learn large margin between positive and negative pairs, we propose a novel multi-stage training strategy. Firstly, we train the aforementioned triplet loss $\mathcal{L}_0$ with a small base margin $m_0$ and obtain a base feature embedding $f_0(x_i)\in\mathbb{R}^{d}$. Meanwhile, we learn a feature shift vector $f^s_{1}(x_i)$ which is of the same dimension as $f_0(x_i)$. By adding the two vectors together, we get a shifted feature embedding $f_1(x_i)$. This process can be recursively applied by:
\begin{equation}
  f_j(x_i) = f_{j-1}(x_i) + f^s_{j}(x_i), \quad \forall j \ge 1
	\label{shift}
\end{equation}
The feature shifts $f^s_j(x_i)$ are not learned directly. Instead, we supervise the shifted features $f_j(x_i)$ with another triplet loss $\mathcal{L}_j$. To make sure that each time of feature shifting results in better feature embedding, the margin of $\mathcal{L}_j$ is monotonically increased by $m_j = m_{j-1} + \Delta m_j$. The $j$-th triplet loss is defined as:
\begin{equation}
  \begin{aligned}
    \mathcal{L}_{j} &= \sum_i^N\Big[d^{ap}_j - d^{an}_j + m_{j}\Big]_+, \\
        d^{ap}_j &= ||f_j(x_i^a) - f_j(x_i^p)||_2^2 \\
        d^{an}_j &= ||f_j(x_i^a) - f_j(x_i^n)||_2^2 
  \end{aligned}
  \label{l-shift}
\end{equation}
The incremental design makes the learning of large margin easier. The gap between distance of negative pairs $d^{an}$ and that of positive pairs $d^{ap}$ gets enlarged progressively. And we empirically demonstrate that larger margin trained by this way leads to better performance.

All of the triplet losses at different stages are optimized jointly. The final loss is the weighted sum of all losses:
\begin{equation}
  \mathcal{L} = \sum_{j=0}^M \lambda_j\mathcal{L}_j 
  \label{l-final}
\end{equation}
where $M$ is the number of times that feature shifting is applied. $\lambda_j$ is the weight used to balance different losses. In our experiments, we set $M$ to 2 and $\lambda_j$ to 1 in all stages.

\subsection{Exploiting Multiple Levels of Features}
\begin{figure*}[t]
	\centering
	\includegraphics[width=.7\linewidth]{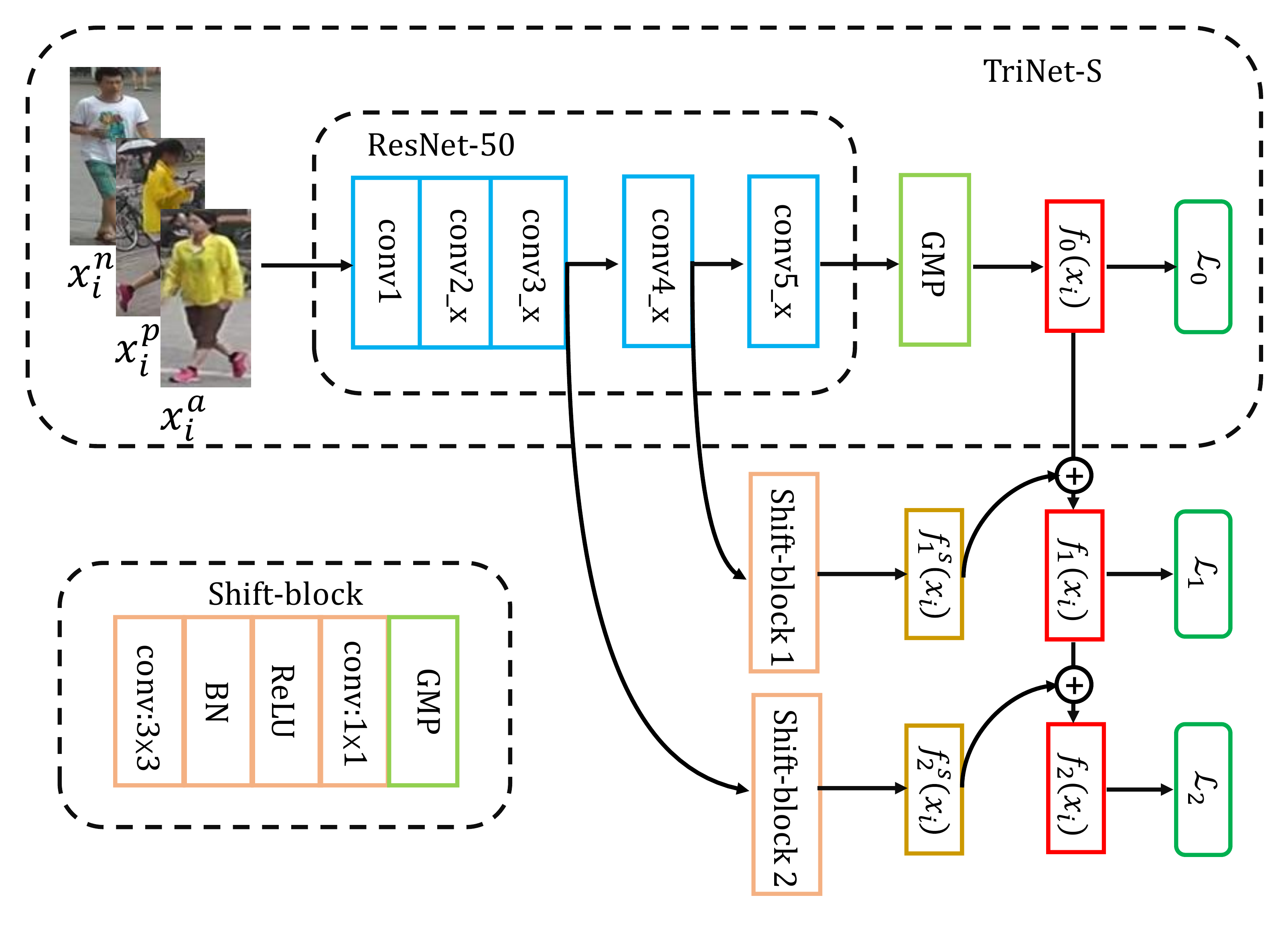}
    \caption{Pipeline of the proposed LITM approach. Upper: a strong triplet loss baseline network (TriNet-S). Lower: learning feature shift vectors for recursively shifted features.}
	\label{network}
\end{figure*}

High-level feature maps of a neural network contain coarse semantic information, while mid-level features contain detailed structure information. In previous works, only high-level features have been exploited. We argue that mid-level features are important for fine-grained visual recognition tasks like person ReID. Our multi-stage framework easily enjoys the benefits of different levels of features. Specifically, we learn the base feature embedding $f_0(\cdot)$ using high-level features, which serves as a decent starting point. Then mid-level features are exploited to learn the feature shifts $f_j^s(\cdot)$, which requires a closer look at subtle appearance differences between two persons. See Figure~\ref{network} for the pipeline of our framework.

\subsection{Global Hard Identity Searching}

\begin{algorithm}[t]
	\caption{Global Hard Identity Searching.}
	\label{alg-ghis}
	\begin{algorithmic}[1]
		\Require Training set of $n$ identities, feature extractor
		\Ensure Hard identity sets of all $n$ identities $S$
		\State Compute the mean distance matrix $\bar D$ with Equation~\eqref{mean-distance}
        \State Set diagonal elements $\{D_{u,u}\}$ in $\bar D$ to infinity
        \For {each identity $u=1, \dotsc, n$}
        \State Find the $g$ most similar candidate identities $C_u$ according to $\bar D_{u,\ast}$
		\State Generate hard identity set $S_u$ by randomly sampling $q$ identities from $C_u$
        \EndFor \\
        \Return $S = \{S_1, \dotsc, S_n\}$
	\end{algorithmic}
\end{algorithm}

To produce triplets with high quality negative pairs, we introduce a global hard identity searching method. When generating a training batch, \cite{hermans2017defense} sample $P$ identities randomly. Instead we define a metric to measure the dissimilarity between two identities and put similar identities together in the same batch. Given a training set with $n$ identities, we randomly sample $K$ examples per-identity. Then we compute the pairwise mean distance matrix $\bar D$ of the $n$ identities. $\bar D$ is an $n \times n$ symmetric matrix. $\bar D_{u,v}$ measures the dissimilarity between identity $u$ and $v$, which is defined as:
\begin{equation}
  \bar D_{u,v} = \frac{1}{K^2}\sum_{l=1}^{K}\sum_{r=1}^{K}||f_M(x^u_l) - f_M(x^v_r)||_2^2
  \label{mean-distance}
\end{equation}

The searching procedure is described in Algorithm~\ref{alg-ghis}. After computing the mean distance matrix $\bar D$, we set its diagonal elements $\{\bar D_{u,u}\}$ to infinity to prevent each identity itself from being sampled as its hard identity. Then for each identity $u$, we find $g$ identities of the smallest distances to $u$ as its candidate hard identities. To introduce more randomness, $q$ ($q < g$) identities are sampled from the $g$ candidates as the final hard identity set. Each identity and its $q$ hard identities form an identity group which contains $q+1$ identities. When generating a training batch, we sample $P / (q+1)$ identity groups, which results in $P$ identities in total. After a training batch is generated, we use batch hard triplet mining \cite{hermans2017defense} to sample hard triplets within the batch.

In our experiments, we notice that the hard identity set of a identity rarely changes during training. To cover more identity permutations and make the training more stable, we apply GHIS and random identity sampling in an alternating way. We first train the network with random identity sampling for two epochs, followed by GHIS for one epoch. This procedure is repeatedly applied. As for other hyper-parameters, we set $g=5$, $q=3$, $P=20$ and $K=4$.

\subsection{TriNet-S: A Strong Triplet Loss Baseline Network}
For the network architecture, we consider TriNet proposed in \cite{hermans2017defense} as a reference. TriNet is adapted from ResNet-50 \cite{he2016deep}, where the last fully connected layer is replaced with two new fully connected layers. The first layer reduces the feature dimension from 2048 to 1024. And the second layer further reduces the dimension to 128, which serves as final feature embedding. We argue that this configuration is not optimal. For person ReID which involves fine-grained recognition, we propose the following guidelines on network design.
\begin{itemize}
  \item A fully convolutional architecture is preferable to learn spatial-aware features.
  \item Global maximum pooling results in sharper and thus more discriminative responses than global average pooling.
  \item Resolution matters. Large feature map size is preferable as more detailed information is preserved.
\end{itemize}

Following these guidelines, we make some tweaks to TriNet. Firstly, we remove the last two fully connected layers and use the globally pooled features as final feature embedding. Secondly, we replace the global average pooling (GAP) of ResNet-50 with global maximum pooling (GMP). Thirdly, we reduce the stride of the first convolutional layer in the conv5\_x block from 2 to 1, which doubles the feature map size. With these tweaks alone, we achieve significant performance gain. We refer to our implementation as TriNet-S, which serves as a strong triplet loss baseline.

\subsection{Network Architecture}
Combining TriNet-S and LITM, our network architecture is shown in Figure~\ref{network}. Following \cite{hermans2017defense}, we use ResNet-50 as our backbone network. The 2048-dimensional output of GMP is utilized as the base feature embedding $f_0(\cdot)$. The feature maps of conv4\_x and conv3\_x are fed into two shift blocks respectively, producing two feature shift vectors: $f^s_1(\cdot) $ and $ f^s_2(\cdot)$. Then the shifted features are created by adding the shift vector to the base feature vector. The shift block is a tiny sub-network as shown in Figure~\ref{network}. The number of channels of the first $3 \times 3$ convolution is kept the same as its input channels (i.e. 1024 for shift-block1 and 512 for shift-block2). The second $1 \times 1$ convolution is utilized to increase the feature dimension to 2048. Notably, feature maps in shift-block2 are down-sampled by half by setting stride of the first convolution to 2. During training, all the three triplet losses are optimized jointly. During inference, only the final shifted feature embedding $f_2(\cdot)$ is used.

\section{Experiments}\label{experiments}
\subsection{Datasets}
We evaluate the proposed approach on three large-scale person ReID datasets, namely Market-1501 \cite{zheng2015scalable}, CUHK03 \cite{li2014deepreid} and DukeMTMC-reID \cite{ristani2016performance,Zheng_2017_ICCV}.

\begin{itemize}
	\item \textbf{Market-1501} contains altogether 32,688 images of 1,501 labeled pedestrians, which were captured under 6 camera viewpoints in a campus. Deformable Part Model (DPM) \cite{felzenszwalb2008discriminatively} is employed to produce pedestrian bounding boxes. This dataset is split into two non-overlapping partitions: 12,936 images from 751 identities (including 1 background category) for training and 19,732 images from 750 identities for testing. During testing, 3,368 images are chosen as query images. We adopt single-query evaluation mode in all experiments.
	
	\item \textbf{CUHK03} contains 14,096 pedestrian images of 1,467 identities. Each person image in this dataset was captured from two different cameras in the CUHK campus. It provides both DPM-detected and hand-marked bounding boxes. In this paper, we report experimental results on both image sets. We utilize the more challenging train/test split protocol proposed in \cite{Zhong_2017_CVPR} where 767 identities are used for training and the rest 700 for testing.
	
	\item \textbf{DukeMTMC-reID} is a subset of Duke-MTMC for ReID. The images were captured with 8 cameras for cross-camera tracking. It contains 16,522 training images from 702 identities, 2,228 queries from the other 702 identities and 17,661 gallery images. On this dataset, we also test our method in the single-query setting.
\end{itemize}

\subsection{Evaluation Metrics}
Following most existing person ReID works, we use two evaluation metrics to evaluate the performance of our method. One is the Cumulated Matching Characteristics (CMC), which considers ReID as a ranking problem. The other is mean average precision (mAP), which considers ReID as a retrieval problem.

\subsection{Implementation Details}
Our implementation is based on PyTorch \cite{paszke2017automatic}. The backbone ResNet-50 is pre-trained on ImageNet~\cite{ILSVRC15}. We use the same data augmentation across all experiments and on all datasets unless otherwise noted. The training images are randomly cropped with a ratio uniformly sampled from $[0.8, 1)$ and resized to $288 \times 144$. Random erasing \cite{Zhong2017Random} and random flipping are applied on resized images with a probability of $0.5$. The hyper-parameters of random erasing data augmentation are set the same as \cite{Zhong2017Random}. The number of persons $P$ per-batch and number of images per-person $K$ are set to 20 and 4 respectively. Hence, the mini-batch size is 80. For LITM, the base and incremental margins are set as $m_0 = 4$, $m_1 = 7$, $m_2 = 10$.

We use the Adam optimizer \cite{Kingma2014Adam} with $\epsilon=10^{-3}$, $\beta_1=0.99$ and $\beta_2=0.999$. The network is trained for 300 epochs in total. And a piecewise learning rate schedule is utilized, where it is fixed to $2\times 10^{-4}$ in the first 150 epochs and decayed exponentially in the rest 150 epochs.
\[
  lr(t) = 
  \begin{cases}
    2 \times 10^{-4} & \quad \text{if } t \leq 150 \\
    2 \times 10^{-4} \times 10^{-3\times\frac{t-150}{150}} & \quad \text{if } 150 < t \leq 300
  \end{cases}
    \label{lr-schedule}
\]

\subsection{Improvements over Triplet Loss Baseline}
We first report the performance gains brought by our tweaks to the network architecture in TriNet-S on the Market-1501 dataset. As shown in Table~\ref{cmp-trinet}, after removing the trailing fully connected layers and making the network fully convolutional, mAP gets improved by 1\% from 70.6\% to 71.6\%. Replacing GAP with GMP brings more than 4\% performance gains in terms of both Rank-1 accuracy and mAP. By reducing the stride of the conv5\_x block from 2 to 1 and thus increasing the feature map resolution, we obtain an extra 2\% mAP gain. Compared with the TriNet baseline, the proposed TriNet-S improves mAP by 7.3\% and Rank-1 accuracy by 4.8\%.
\begin{table}[t]
  \centering
  \begin{tabular}{c c c|c c} 
    \hline
    Stride  & Pooling & Fully Conv. & Rank-1 & mAP \\
    \hline  
    2 & GAP &  & 85.3 & 70.6 \\
    \hline
    2 & GAP & \checkmark & 85.4 & 71.6  \\
    2 & GMP & \checkmark & 89.7 & 75.9 \\
    1 & GMP & \checkmark & 90.1 & 77.9\\
    \hline
  \end{tabular}
  \caption{Performance improvements of the proposed TriNet-S over the TriNet baseline on the Market-1501 dataset. The performance of TriNet in this table is slightly better than that reported in \cite{hermans2017defense} because of the random erasing data augmentation we adopt.}
  \label{cmp-trinet}
\end{table}

The proposed LITM training strategy is agnostic to the choice of network architecture. To validate the effectiveness of LITM, we apply it to both TriNet and TriNet-S. As shown in Table~\ref{cmp-to-triloss}, LITM is able to improve TriNet by 6.3\% and 3.1\% in terms of mAP and Rank-1 accuracy respectively on the Market-1501 dataset. Even though TriNet-S have already greatly improved the performance over the TriNet baseline, LITM still boosts mAP by 4.4\% and Rank-1 accuracy by 2.5\%. Similar performance boosts are observed on the CUHK03 and DukeMTMC-reID datasets, which indicates that our approach generalizes well across different scenarios.
\begin{table*}[t]
  \centering
  \begin{tabular}{c|c|c c|c c|c c} 
    \hline
    \multirow{2}{*}{Network} & \multirow{2}{*}{LITM} & \multicolumn{2}{|c|}{Market-1501}&\multicolumn{2}{|c|}{CUHK03 (labeled)}&\multicolumn{2}{|c}{DukeMTMC-reID}\\
    & & Rank-1 & mAP & Rank-1 & mAP & Rank-1 & mAP \\
    \hline  
    TriNet~\cite{hermans2017defense} &  &84.9&69.1&55.2$^\dagger$&54.3$^\dagger$&76.4$^\dagger$&60.4$^\dagger$\\
    TriNet~\cite{hermans2017defense} & \checkmark &88.0 &75.4 &60.9  &59.3  &79.2 &65.9 \\
    \hline
    TriNet-S &  &90.1&77.9&63.5&61.7&82.8&70.2\\
    TriNet-S & \checkmark &\textbf{92.6}&\textbf{82.3}&\textbf{73.1}&\textbf{71.0}&\textbf{84.8}&\textbf{74.4}\\
    \hline
  \end{tabular}
  \caption{Performance improvements of LITM to both TriNet and TriNet-S. $^\dagger$ indicates reproduced results by us using the same training configuration as the TriNet paper.}
  \label{cmp-to-triloss}
\end{table*}

\subsection{Comparisons with the State-of-the-arts}

\subsubsection{Results on Market-1501}
As shown in Table~\ref{cmp-market1501}, GHIS further brings 1.3\% and 1.6\% improvements for Rank-1 accuracy and mAP respectively. Compared with 11 recently proposed methods on the Market-1501 dataset, our final result yields the best mAP (83.9\%) and comparable Rank-1 accuracy (93.9\%) to SphereReID. Although SphereReID achieves the best Rank-1 accuracy, its optimization is very sensitive to hyper-parameter settings. For example, a carefully designed learning rate warming up schedule is required. In GSRW, testing images are fed into the network in a pairwise manner, which is much more time consuming than our approach. PCB+RPP is trained with a three-stage process with fine-tuning, which is not an end-to-end method.
\begin{table}[t]
	\centering
	\begin{tabular}{l|c c} 
		\hline
        Measure (\%)&Rank-1&mAP\\
		\hline  
		Pose-transfer~\cite{Liu_2018_CVPR}&87.7&68.9\\
		AOS~\cite{Huang_2018_CVPR}&86.5&70.4\\
		MGCAM~\cite{Song_2018_CVPR}&83.8&74.3\\
		MLFN~(Chang et al. 2018)&90.0&74.3\\
		HA-CNN~\cite{Li_2018_CVPR}&91.2&75.7\\
		AlignedReID$^*$~\cite{Zhang2017AlignedReID}&91.8&79.3\\
		Deep-Person$^*$~\cite{Bai2017Deep}&92.3&79.6\\
		GCSL~\cite{Chen_2018_CVPR}&93.5&81.6\\
		PCB+RPP$^*$~\cite{Sun2018Beyond}&93.8&81.6\\
		GSRW~\cite{Shen2_2018_CVPR}&92.7&82.5\\
		SphereReID$^*$~\cite{Fan2018SphereReID}&\textbf{94.4}&83.6\\
		\hline
		LITM&92.6&82.3\\
		LITM+GHIS&93.9&\textbf{83.9}\\
		\hline
	\end{tabular}
	\caption{Performance comparison on the Market-1501 dataset. $^*$~denotes unpublished work on arXiv.}
	\label{cmp-market1501}
\end{table}

\subsubsection{Results on CUHK03}
We choose the new training/testing split protocol proposed in \cite{Zhong_2017_CVPR} instead of the original protocol for convenience. A comparison our approach with recent methods following the same evaluation protocol are listed in Table~\ref{cmp-cuhk03}. LITM+GHIS outperforms the $2^{\rm nd}$ best approach (PCB+RPP) by 8.1\% (71.8\% vs. 63.7\%) for Rank-1 accuracy and 11.6\% (69.1\% vs. 57.5\%) for mAP. The significant performance advantage fully validates the superiority of the proposed LITM and GHIS over existing methods.
\begin{table}[t]
	\centering
		\begin{tabular}[width=.7\textwidth]{l|c c|c c} 
			\hline
			Data Type &\multicolumn{2}{|c|}{Labeled} &\multicolumn{2}{|c}{Detected} \\
			Measure (\%)&Rank-1&mAP&Rank-1&mAP\\
			\hline 
			HA-CNN~\shortcite{Li_2018_CVPR}&44.4&41.0&41.7&38.6\\
			Pose-transfer~\shortcite{Liu_2018_CVPR}&45.1 &42.0 &41.6 &38.7\\
			MGCAM~\shortcite{Song_2018_CVPR}&50.1&50.2&46.7&46.9\\
			AOS~\shortcite{Huang_2018_CVPR}&-&-&47.1&43.3\\
			MLFN~\shortcite{Chang_2018_CVPR}&54.7&49.2&52.8&47.8\\
			REDA$^*$~\shortcite{Zhong2017Random}&58.1 &53.8 &55.5 &50.7\\
			PCB+RPP$^*$~\shortcite{Sun2018Beyond}&-&-&63.7&57.5\\
			\hline
			LITM&73.1&71.0&71.0&68.6\\
			LITM+GHIS&\textbf{74.2}&\textbf{71.7}&\textbf{71.8}&\textbf{69.1}\\
			\hline
		\end{tabular}
	\caption{Performance comparison on the CUHK03 dataset. $^*$~denotes unpublished work on arXiv.}
	\label{cmp-cuhk03}
\end{table}

\subsubsection{Results on DukeMTMC-reID} Compared with Market-1501, pedestrian images from this dataset have more variations in illumination and background because of wider camera views and more complex scene layout. On this challenging dataset, our LITM+GHIS approach again outperforms all recent methods by a large margin as shown in Table~\ref{cmp-duke}. Notably, our approach outperforms SphereReID \cite{Fan2018SphereReID} by 2.0\% and 6.0\% in terms of Rank-1 accuracy and mAP respectively, which indicates that our improvements on training strategy and network architecture are general and work well in a wide variety of scenarios.

\begin{table}[t]
	\centering
	\begin{tabular}{l|c c} 
		\hline
		Measure (\%)&Rank-1&mAP\\
		\hline  
		Pose-transfer~\cite{Liu_2018_CVPR}&78.5&56.9\\
		AOS~\cite{Huang_2018_CVPR}&79.2&62.1\\
		MLFN(Chang et al. 2018)&81.0&62.8\\
		HA-CNN~\cite{Li_2018_CVPR}&80.5&63.8\\
		Deep-Person$^*$~\cite{Bai2017Deep}&80.9&64.8\\
		GSRW~\cite{Shen2_2018_CVPR}&80.7&66.4\\
		SphereReID$^*$~\cite{Fan2018SphereReID}&83.9&68.5\\
		PCB+RPP$^*$~\cite{Sun2018Beyond}&83.3&69.2\\
		GCSL~\cite{Chen_2018_CVPR}&84.9&69.5\\
		\hline
		LITM&84.8&74.4\\
		LITM+GHIS&\textbf{85.9}&\textbf{74.5}\\
		\hline
	\end{tabular}
	\caption{Performance comparison on the DukeMTMC-reID dataset. $^*$~denotes unpublished work on arXiv.}
	\label{cmp-duke}
\end{table}

\section{Ablation Studies}
To further investigate the design choices of our approach, we perform extensive ablation studies on the Market-1501 dataset. In particular, we compare the behaviors of GAP and GMP, the impact of incremental triplet margin and alternative LITM structures.

\subsection{GAP vs. GMP} By analyzing the feature maps before global pooling, we find that the great majority of elements are close to 0. Therefore, the average operation in GAP would greatly reduce the magnitude of feature vectors, which weakens the feature discriminativeness. To prove the hypothesis, we compute the mean distance of positive and negative pairs after the training converges. Table~\ref{cmp-distance} shows a comparison of GAP and GMP in terms of the mean distance.
\begin{table}[t]
	\centering
	\begin{tabular}{l|c c | c} 
		\hline
		Pooling &$\bar{d}^{ap}$&$\bar{d}^{an}$&$\bar{d}^{an}-\bar{d}^{ap}$\\
		\hline  
		GAP&10.5&25.1&14.6\\
		GMP&40.9&66.2&25.3\\
		\hline
	\end{tabular}
	\caption{The mean distance of positive and negative pairs regarding different global pooling methods in TriNet.}
	\label{cmp-distance}
\end{table}
By replacing GAP with GMP, mean distance of both positive and negative pairs gets significantly increased. At the same time, although trained with the same triplet margin, the gap between $\bar{d}^{an}$ and $\bar{d}^{ap}$ is enlarged from 14.6 to 25.3. And the performance also gets significantly improved as shown in Table~\ref{cmp-gmp}. This again verifies that larger margin leads to better feature embedding.
\begin{table}[t]
	\centering
	\begin{tabular}{l| c |c c} 
		\hline
		Method  &pool  &Rank-1&mAP\\
		\hline  
		\multirow{2}{*}{TriNet}  &GAP  &85.4 &71.6  \\
		  &GMP  &89.7 &75.9 \\
		 \hline  
		\multirow{2}{*}{LITM}  &GAP &89.3 &77.9  \\
		     &GMP   &92.6 &82.3\\
		\hline
	\end{tabular}
	\caption{Performance improvements of GMP over GAP on the Market-1501 dataset.}
	\label{cmp-gmp}
\end{table}
\subsection{Impact of Incremental Margin} To validate that the multi-stage triplet losses with incremental margins have learned increasingly better feature embedding, we first compare the mean distance of positive and negative pairs at different stages. As show in Table~\ref{litm-distance}, from $f_0(\cdot)$ to $f_2(\cdot)$ the distance of positive pairs $\bar{d}^{ap}$ and negative pairs $\bar{d}^{ap}$, as well as their gap are progressively increased. In particular, the distance gap between positive and negative pairs increases from 25.9 to 32.7. In terms of performance, as shown in Table~\ref{cmp-diff-delta}, shifted features are also superior over base features. With a single iteration of feature shifting, $f_1(\cdot)$ boosts mAP from 80.9\% to 82.2\%. While improvement from more iterations is marginal.
\begin{table}[t]
	\centering
	\begin{tabular}{l|c c | c} 
		\hline
		Feature &$\bar{d}^{ap}$&$\bar{d}^{an}$&$\bar{d}^{an}-\bar{d}^{ap}$\\
		\hline  
		$f_0(\cdot)$&59.0&84.9&25.9\\
		$f_1(\cdot)$ &68.9&99.1&30.2\\
		$f_2(\cdot)$&70.9&103.6&32.7\\
		\hline
	\end{tabular}
	\caption{The mean distance of positive and negative pairs of features at different stages of LITM.}
	\label{litm-distance}
\end{table}

To validate the effectiveness of our incremental triplet margin strategy, we compare it with the one-stage large-margin triplet loss. Specifically, we train TriNet-S with different margins. From Table~\ref{cmp-diff-delta}, we can see that performance of TriNet-S gets improved when the margin increases from 1 to 4, but degrades for larger margins. Notably, the performances of $f_0(\cdot)$, $f_1(\cdot)$ and $f_2(\cdot)$ consistently outperform their TriNet-S counterparts with the same margin value, which clearly demonstrates the superiority of the proposed LITM method.
\begin{table}[t]
	\centering
	\begin{tabular}{c| c |c c c c} 
		\hline
		 & $m$ &Rank-1&Rank-5&Rank-10&mAP\\
		\hline  
		\multirow{4}{*}{TriNet-S}&1&90.1&94.8&96.5&77.9\\
		 &4 &90.9&96.6&97.5&79.1\\
		 &7 &90.2&96.4&97.0&78.2\\	
		 &10 &89.9&95.2&96.2&77.7\\	
		\hline  
		$f_0(\cdot)$ &4&92.1&96.9&98.0&80.9\\
		$f_1(\cdot)$ &7&92.6&97.1&98.5&82.2\\
		$f_2(\cdot)$ &10&92.6&97.5&98.5&82.3\\
		\hline
	\end{tabular}
	\caption{Performance of feature embeddings at different stages of LITM and comparison with TriNet-S with various margins on the Market-1501 dataset.}
	\label{cmp-diff-delta}
\end{table}

\subsection{Alternative LITM Structures}
We further compare the current LITM structure in Figure~\ref{network} with two alternatives.
\begin{itemize}
  \item LITM-C5C5C5: the base features as well as shifted features are learned from the conv5\_x block.
  \item LITM-C3C4C5: the base features and shifted features are learned from the conv3\_x, conv4\_x and conv5\_x blocks respectively.
  \item LITM-C5C4C3: the current structure we use in our experiments.
\end{itemize}
Table~\ref{cmp-diff-stru} shows the results. The C5C4C3 setting outperforms C5C5C5, which validates that mid-level features indeed help. While C3C4C5 is the worst. The reason is that a decent base feature embedding learned from high-level feature maps is critical.
\begin{table}[t!]
	\centering
	\begin{tabular}{c|c c c|c} 
		\hline
		Measure (\%)&Rank-1&Rank-5&Rank-10&mAP\\
		\hline  
		LITM-C5C5C5&92.0&97.0&98.2&81.2\\
		LITM-C3C4C5&90.8&96.3&97.9&79.4\\
		LITM-C5C4C3&92.6&97.5&98.5&82.3\\
		\hline
	\end{tabular}
	\caption{Performance comparison of different LITM structures on the Market-1501 dataset.}
	\label{cmp-diff-stru}
\end{table}

\section{Conclusion}\label{conclusion}
In this paper, we verify that triplet loss is an effective tool to learn discriminative features for person ReID. However, existing training framework is far from optimal. By learning incremental triplet margin, global hard identity searching and a better network architecture, we make significant performance improvement and achieve state-of-the-art performances on common person ReID datasets. Our improvements to triplet loss may also apply to other related visual tasks, such as face recognition and object retrieval. We leave this as future work.

\bibliography{e2h-reference} 
\bibliographystyle{aaai}

\end{document}